\documentclass[lettersize,journal]{IEEEtran}

\usepackage{graphicx, subfigure}
\usepackage{amssymb}
\usepackage{enumitem}
\usepackage{listings}
\usepackage{tabularx}
\usepackage{xcolor}
\usepackage[boxed]{algorithm}
\usepackage{algorithmicx}
\usepackage{url}
\usepackage{colortbl}\usepackage{tabu}
\usepackage{array, ragged2e}
\usepackage{graphicx,epsfig,epstopdf,amsmath,amssymb,enumerate} 
\usepackage{algpseudocode}
\usepackage{hyperref}
\usepackage{xspace}
\usepackage{color,soul}
\usepackage[colorinlistoftodos]{todonotes}
\usepackage[numbers,sort&compress]{natbib}

\begin{document}

\title{SeaNet: Towards A Knowledge Graph Based Autonomic Management of Software Defined Networks}

\author{Qianru~Zhou~\IEEEmembership{IEEE Member}, %
        Alasdair J. G. Gray, 
        Stephen~McLaughlin$^\dag$~\IEEEmembership{IEEE Fellow}
\thanks{This research was supported by the EPSRC TOUCAN project (Grant No. EP/L020009/1), National Natural Science Foundation of China (No. 61973161, 61991404), Future network research fund project (FNSRFP-2021-YB-05), and the Starting Program of Nanjing
University of Science and Technology (No. AE89991/324).}%
\thanks{Qianru Zhou is with the School of Computer Science and Engineering, Nanjing University of Science and Technology, Nanjing, P.R.China. e-mail: (zhouqianru@njust.edu.cn).}
\thanks{Alasdair J. G. Gray is with the Department of Computer Science, Heriot-Watt University, Edinburgh, EH14 4AS, U.K.}
\thanks{Stephen McLaughlin is with the School of Engineering \& Physical Sciences, Heriot-Watt University, Edinburgh, EH14 4AS, U.K.}
\thanks{$^\dag$ corresponding author}

}

\markboth{Journal of \LaTeX\ Class Files,~Vol.~14, No.~8, August~2021}%
{Shell \MakeLowercase{\textit{et al.}}: A Sample Article Using IEEEtran.cls for IEEE Journals}

\IEEEoverridecommandlockouts
\IEEEpubid{\makebox[\columnwidth]{0000--0000/00\$00.00~\copyright~2021 IEEE \hfill} \hspace{\columnsep}\makebox[\columnwidth]{ }}
\maketitle
\IEEEpubidadjcol


\begin{abstract}
Automatic network management driven by Artificial Intelligent technologies is quite pervasive today. However, most available reports focus on theoretic proposals and architecture designs; works with details of practical implementations of ontological autonomic networks management are yet to appear. In this paper we presents our effort of an knowledge graph-driven methodology for autonomic network management in software defined networks (SDNs), termed as SeaNet. Driven by the ToCo ontology, SeaNet is developed based on an open-source SDN emulator -- Mininet. There are three core components: a \emph{knowledge graph generator}, a \emph{SPARQL engine}, and a \emph{network management API}. The \emph{knowledge graph generator} represents the knowledge in the telecommunication network resources and management tasks into a knowledge base using formal ontology. Expert experience and network management rules can also be formalized into knowledge graph. By automatically inference by the \emph{SPARQL engine}, \emph{the network management API} is able to packet technology-specific details and expose technology-independent interfaces to users. We designed and carried out a series of experiments to compare SeaNet with a commercial SDN controller Ryu implemented by the same programming language Python. The evaluation results show that SeaNet is considerably faster in most circumstances than Ryu and the SeaNet code is significantly more compact. Benefit from resource description framework (RDF) reasoning, SeaNet is able to achieve $O(1)$ time complexity on different scales of the knowledge graph while the traditional database can achieve $O(nlogn)$ at its best. With the network management API, SeaNet enables researchers to develop semantic-intelligent applications on their own SDNs.

\end{abstract}


\section{Introduction}\label{s1}
\IEEEPARstart{T}{he} rapid expansion of Internet services has resulted in today's network management tasks becoming increasingly complex \cite{houidi2016knowledge}. Despite heated discussions over developing autonomic network management in the community for decades, the current network management system remains to spend significant efforts on low-level, tedious, error-prone tasks and often requires complex manual reasoning from network administrators \cite{houidi2016knowledge,knownet2016quinn,ravel2016wang,human2018rojas,knowledge2017mestres,Barakat2017GavelSN}. As noted by Shenker \emph{et. al}, the ultimate goal of a software defined network (SDN) is to find appropriate abstractions that enable the complex network management tasks to be broken into tractable elements \cite{1}. More and more researchers believe a knowledge base that can capture global knowledge from networks and allow the management system to readily reason over it is the answer to Shenker's quesiton \cite{human2018rojas,knowledge2017mestres}. 

Coined by Google in 2012, the term \emph{knowledge graph} is to name Google's own knowledge base, proposed to enhance its search results with linked information gathered from a variety of sources. With the success of Google, the term ``knowledge graph'' is way more popular than ``knowledge base'' (which is what it really is), we will use both ``knowledge graph'' and ``knowledge base'' alternatively in the following part of the paper, where no ambiguity will occur.  

Rooted on Artificial Intelligent, knowledge graph has gained more and more attention in network management \cite{knownet2016quinn,human2018rojas,knowledge2017mestres}. Driven by ontology, knowledge graph is able to extract from logs, records, and documents, structured or not, to formal and consensual terminologies and link them together with semantic relations \cite{knownet2016quinn,human2018rojas,knowledge2017mestres}. The core idea of knowledge based autonomic network management systems is to represent all the observations, measurements, concepts, expertise, and the relationships between them using formal logic (e.g., first order logic), and automatically reasoning over it. Besides, driven by ontology, network management rules and policies and human expert experiences can be represented by formal logical languages in a knowledge graph, allowing not only knowledge reuse, but also automatic reasoning over the knowledge base of the whole network, not only part of it, and thus eliminate human intervention \cite{Barakat2017GavelSN,knownet2016quinn,human2018rojas}. 

Although numerous papers have been reported for knowledge graph based network management, most of these works focus on using machine learning only \cite{Barakat2017GavelSN,knownet2016quinn,human2018rojas}, even for those works that involve knowledge based technologies mainly focus on theoretical proposals or architecture designs only \cite{knowledge2017mestres}, and very few reports practical experiments on real-life network testbed. In this paper, we take a modest step towards implementing a knowledge graph based autonomic management system. We propose \textbf{SeaNet} (\textbf{S}emantic \textbf{E}nabled \textbf{A}utonomic Management of Software Defined \textbf{Net}works), and evaluate its practicability and efficiency. We demonstrate how it provides the capability to 1) abstract the network information into an ontology-assisted knowledge base, 2) interpret higher-level requirements into lower-level operations, and 3) execute the operations automatically. SeaNet system consists of three parts: a knowledge base generator, a SPARQL engine, and a network management API. Evaluation is carried out by testing some use cases of basic network management task through the open SeaNet API. The performance of SeaNet is compared with Ryu, an SDN controller developed with the same language Python as SeaNet. 

The rest of the paper is organized as follows. Section \ref{sec_related} discusses the background and requirements for the research on knowledge based autonomic network management. The architecture of SeaNet and the technologies adopted are presented in Section \ref{sec_system}, together with details of ontology used in SeaNet in Section \ref{sec_onto}. Section \ref{sec_demo} evaluates the practicability of each component of the SeaNet system, including modeling and reasoning ability. The performance of the API methods is compared with the SDN controller Ryu. Finally, conclusions are drawn in Section \ref{sec_conclusion_section}.

\section{State of the art}\label{sec_related}

Investigation on using artificial intelligence (AI) to achieve autonomic telecommunication network management has started since the very beginning of artificial intelligence. The first real-time AI system, Interactive Real-Time Telecommunications Network Management System (IRTNMS), was developed by Australia Artificial Intelligence Institute (AAII) and Telecom Australia in 1992 \cite{ingrand1992architecture}. Based on Procedural Reasoning Systems (PRS), IRTNMS implements BDI (Belief--Desire--Intention) architecture to diagnose, control, and monitor the telecommunications network. LODES (Large-intenetwork Observation and Diagnostic Expert System) is an expert system for detecting and diagnosing problems in a segment of a local area network \cite{sugawara1992multiagent}.

Recent implementations include Experiential Networked Intelligence Industry Specification Group (ENI ISG), which aims at assisting 5G network deployment and operation with closed-loop AI mechanisms based on context-aware, metadata-driven policies. ENI ISG proposed a Cognitive Network Management architecture in 2017, using AI techniques and context-aware policies to adjust offered services based on dynamic user needs, environmental conditions and business goals.

With the network management tasks growing complicated, more sophisticated AI technologies are deployed. KnowNet \cite{knownet2016quinn} is inspired from NetSearch, a search based autonomic network management, using information retrieval technologies to accomplish a ``Google'' for network management. KnowNet claims itself go beyond information retrieval to knowledge graph, and have developed a knowledge graph platform for autonomic network management tasks such as network connecting, conflict resolution, and network security. Use cases are given to evaluate the platform, however, details of the experiment implementation, such as the information retrieval process, knowledge graph construction process and inference process, are missing. Also, the details of the knowledge graph it uses, i.e., "kilo", is missing (In \cite{knownet2016quinn}, the authors also fail to declare whether kilo is an ontology or not, if it is, we think the term ``ontology'' should be used instead of knowledge graph, for these two have totally different definitions in computer science and philosophy). 


Like KnowNet, other knowledge based autonomic network management proposals focus on theoretical definitions and architecture describing, rather than provide details on the implementation process on a real-life network testbed, or provide a concrete software tool. The earlier ontological knowledge based network management systems proposed by \emph{L{\'o}pez de Vergara}, \cite{13}, \emph{Ohshima}, \cite{6}, are also discussed below. \emph{L{\'o}pez de Vergara},\cite{13,kanelopoulos2009ontology}, mentioned the possibility of adopting an ontology to map and merge the current heterogeneous knowledge bases, and provide a universal view of the whole managed system. \emph{Strassner} investigated the possibility of ontology-driven policy based network automatic management system. They proposed a series of autonomic network management systems, e.g., Directory Enabled Networks next generation (DEN-ng) policy model, Policy Continuum, and Context-aware policy model. However, as described in the above references, with the absence of practical approaches, these systems have not been evaluated in real-life scenarios. In \cite{6}, \emph{Ohshima} proposed a network management system with the assistance of an ontology, collecting configuration information from traditional network management databases. However, the work was based on a traditional network.\emph{Houidi} \cite{houidi2016knowledge} explored the practicality of a knowledge based automatic reasoner for network management. He implemented the proposed method in python, using Pyke as the reasoner, and tested the system with the basic network connectivity problem. However, in this work, some complexities are hidden for clarity. Thus, it is still not clear whether the system will work in other more dynamic or complicated situations. \emph{Xiao and Xu} \cite{xiao2006integration} tried to integrate ontologies with policy based approaches to accomplish autonomic network management. They also proposed a possible scenario. However, there are no practical tools ready to implement for the proposed methodology, or any real-life evaluations for the proposed system.

\subsection{SeaNet Architecture}
\label{sec_system}

\begin{figure}[hbt]
	\centering
	\includegraphics[width=.4\textwidth]{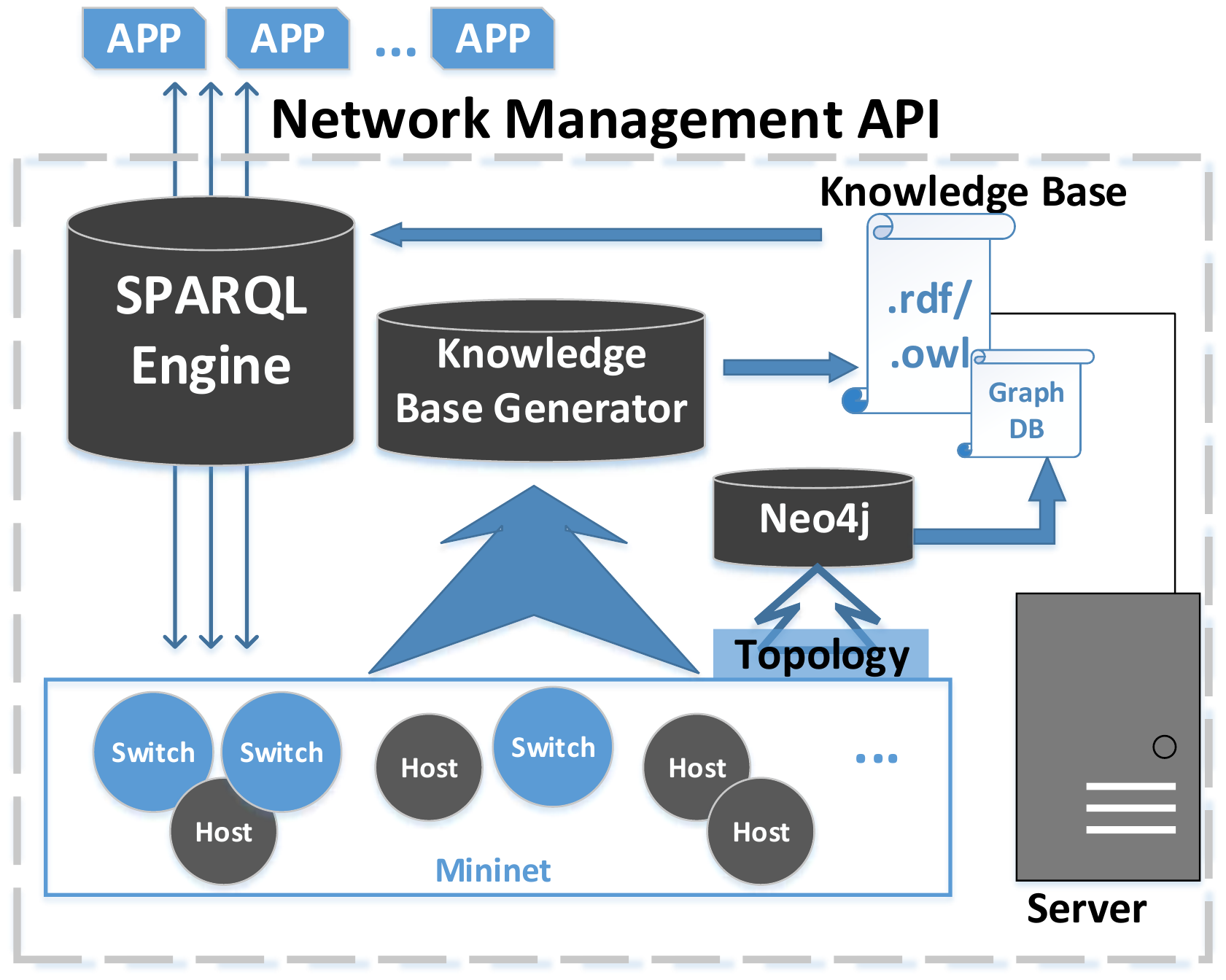}
	\caption{The architecture of proposed knowledge graph based network management system SeaNet.}
	\label{fig_architecture} 
	\vspace*{-\baselineskip}
\end{figure}

In this section, we present the architecture of SeaNet, as shown in Fig. \ref{fig_architecture}. Details of each of the components: knowledge graph constructor, SPARQL engine, and the network management API are discussed. 

The knowledge base generator is designed to bring knowledge base harmonization into reality by retrieving unstructured data from nodes in the network and semantically representing them into the knowledge base following first-order logic format. As shown in Fig. \ref{fig_architecture}, the \emph{knowledge base generator} is designed to integrate data sources from various origins, such as relational database, real-time network measurements and observations, remote data from the cloud and real-time data streams, etc. The global view provided by an ontology is able to achieve direct access to the information from all these sources, and execute inference globally. \emph{Knowledge base generator} provides a novel way to manage and publish network data. With symbolic representation technologies, unstructured and scattered data can be annotated into ontological concepts and ready for reasoning first-order logic formulas.

\subsection{Semantic of SeaNet}

In our prototype SeaNet system, a recursive algorithm is used to read the network records, extract the semantic information, and store it to the knowledge base. In our experiment, when a network has been initiated, the generator can be started by a linux shell command. Information will be collected from network elements into RDF triples. This procedure runs periodically (in this paper, we run it every 5 seconds, but this value is configurable), maintaining an up-to-date knowledge base for the network. A fragment of a knowledge base representing the connectivity and location information of a switch is given below.

\smallskip
\noindent
\texttt{ex:switch1 net:hasPort ex:port1}.\\
\texttt{ex:link1 net:from ex:port1}.\\
\texttt{ex:link1 net:to ex:port2}.\\
\texttt{ex:host1 net:hasPort ex:port2}.\\
\texttt{ex:switch1 geo:location ex:point\_sw1}.\\
\texttt{ex:point\_sw1 geo:long ``123.001''}.\\
\texttt{ex:point\_sw1 geo:lat ``321.001''}.
\smallskip

The \texttt{ex:switch1}, \texttt{ex:port1}, and \texttt{ex:link1} are concepts, and the  \texttt{net:hasPort},  \linebreak\texttt{net:connectTo} are relations. The first four triples denote that \texttt{ex:host1} and \texttt{ex:switch1} are connected by \texttt{ex:link1}. The last three triples denote the location of \texttt{ex:switch1}.  

Semantic queries on the network knowledge base are designed to answer high-level questions such as,  ``Which switch is host1 connected to?'', ``Find me the hosts in the network that are blocked from the others.'' or ``Find me all the hosts connected to switch\_1 and switch\_3, if they are not host\_3 or host\_5,''. With semantic queries, network administration can obtain the knowledge of the network at a highly abstracted level, and thus can issue tasks from an abstract level and leave the detailed technology-specific operations to the SPARQL engine.

To explore SeaNet's ability, APIs are developed to seal all the technology-specific details, and expose only the technology-independent methods, as shown in Table \ref{table_APIsummary}. Although SeaNet is still a prototype system and the APIs presented in this paper could not cover all the network management requirements, we believe these APIs are able to provide automatic mechanisms for some basic, labor intensive, and error-prone network management tasks and bootstrapping. For example, the API method \emph{``add\_flow''} is able to add a flow entry to the flow table in a specific switch, and \emph{``dump\_all\_flows''} can list all the flow entries of the switch. The method \emph{``connectAll''} is able to add connections between all the hosts in a network, forming a mesh network. The method \emph{``buildFirewall''} is design to build a highly customized firewall being able to block dedicated packages between dedicated nodes. All the commands given to these API methods are in plain and simple English, which means customers are not required to have any networking or even computer science background knowledge.

\begin{table*}[ht]
	\caption{Examples of the Network Management API Methods}
	\label{table_APIsummary}
		\begin{tabular}{|p{1.85cm}|p{9cm}|p{5cm}|}
			\hline
			\rowcolor[gray]{.7}
			\Centering\textbf{Functions} & \Centering\textbf{Parameters} & \Centering\textbf{Function}\\ \hline 
			\cellcolor[rgb]{.9,.9,.9}  add\_flow & \emph{\textbf{dst:}} the destination MAC address of the flow entry;\par \emph{\textbf{in\_port:}} the input port id of the flow entry; \par  \emph{\textbf{action\_type:}} action type of the flow entry, could be output, drop, etc.; \par  \emph{\textbf{to\_port:}} the output port id of the flow entry.\par
			& Add a flow entry to the switch's flow table.\\  \hline
			\rowcolor[gray]{.9}
			\cellcolor[rgb]{.9,.9,.9} deleteFlow &  \emph{\textbf{list of input hosts:}} delete the flow from these hosts,\par
			\emph{\textbf{list of output hosts:}} delete the flow & delete a flow from the switch's flow table\\ \hline
			\cellcolor[rgb]{.9,.9,.9} addARPFlow & None & Add an ARP flood flow entry to the switch's flow table, if needed. \\ \hline
			\rowcolor[gray]{.9}
			\cellcolor[rgb]{.9,.9,.9} dump\_all\_flows &  None & Automatically list all the flow entries of the switch. \\ \hline
			
			\cellcolor[rgb]{.9,.9,.9} connectAll &  \emph{\textbf{list of hosts}} & Automatically add flow entries for all the hosts \\ \hline
			\rowcolor[gray]{.9}
			\cellcolor[rgb]{.9,.9,.9} buildFirewall &  \emph{\textbf{list of switches:}} the switches between which the firewall is built; \par  \emph{\textbf{list of hosts:}} the hosts that should be still connected after the firewall is built. & Automatically build a firewall between the given switches, keep the hosts connected in the parameter if given.\\ \hline
			\cellcolor[rgb]{.9,.9,.9} findPath &  \emph{\textbf{list of hosts}} & Automatically find the shortest path between the hosts. \\ 
			\hline
		\end{tabular}
		\vspace*{-\baselineskip}
	\end{table*}
	
	\section{Ontologies in SeaNet}\label{sec_onto}
	
	The ToCo ontology\footnote{\url{http://purl.org/toco/}} proposed for TOUCAN is adopted in SeaNet. ToCo was developed based on the \\ \texttt{Device-Interface-Link (DIL)} Ontology Design Pattern, which was observed and summarized during the ontology engineering process for telecommunication networks with hybrid technologies. Please refer to [32] for details. It was deemed prudent to start from existing available ontologies in the same domain. These were evaluated against our competency questions and where possible reused.
	\par
	For some general resources like units and location, well developed ontologies have been published online, e.g., Unit Ontology (UO)\footnote{http://purl.obolibrary.org/obo/uo.owl} and WGS84 Geo Positioning\footnote{\url{http://www.w3.org/2003/01/geo/wgs84_pos}} are adopted here for units and location resources, respectively. Since ToCo focuses on telecommunication networks, some concepts of the existing network ontology, e.g., Mobile Ontology \cite{178}, are reused. They describe the information of network topology and flows. 
	\par
	As the enabling protocol of SDN, OpenFlow is based on the concept of flows (\texttt{net:Flow}), allowing users to directly control the packet routing inside the network. Each flow is a ``match--action'' pair, which is actually a routing rule, instructing the switch to forward/drop the matched packets. Each switch keeps a collection of flow entries in a flow table. Every time a new packet arrives, the switch searches its flow table for a match (by matching the port (\texttt{net:Interface}) and/or MAC address (\texttt{net:hasMAC})), and executes the corresponding actions (\texttt{net:hasFlowAction}) defined by the flow.
	\par
	Using formal ontology, the system can describe the flows ( \texttt{net:Flow} ) and the actions defined in each flow ( \texttt{net:Action} ). Each \texttt{net:Flow} instance keeps a record of the flow properties such as \texttt{net:flags}, \texttt{net:priority}, \texttt{net:cookie}, \texttt{net:tableId}, \texttt{net:idleTimeout}, and \texttt{net:hardTimeout}. Corresponding flows will be added to the switches through \texttt{net:hasFlow} property. The property \linebreak\texttt{net:toPort} will be used to describe the fact that a flow indicates packets to be forwarded to a port.

\section{Evaluations}\label{sec_demo}
	
	To verify the practicality of SeaNet, the functionality of the system on various networks with different topologies and scales is tested. Mininet is used to networks with different topologies. The whole experimental environment is running in a virtual machine on a MacBook Air, OS X 10.9.5, Intel Core i5, 1.5 GHz, 4GB RAM, Ubuntu 14.04 on VMware workstation. The VMware system has 1GB RAM. The following software tools were used: VirtualBox 5.1.4, rdflib 4.2.1., Neo4j, Open vSwitch (OVS).

\subsection{Evaluations}\label{sec_evaluation}
	
	Evaluations are carried out on computer networks and WiFi networks emulated by Mininet-WiFi, with topologies of linear, single, and tree, and the network scales varying from simple (3 hosts) to massive scale ($\geq 1000$ hosts). The largest network is the one with tree topology, depth = 3, fanout = 5, which contains 1146 nodes (1021 switches and 125 hosts). The smallest network is a single network with $\emph{k} = 5$ (\emph{k} is the number of hosts connected to each switch), which has 6 nodes (1 switch and 5 hosts) in total.
	
\begin{figure}[h]
\vspace*{-\baselineskip}
\begin{center}
	\includegraphics[width=3.3in]{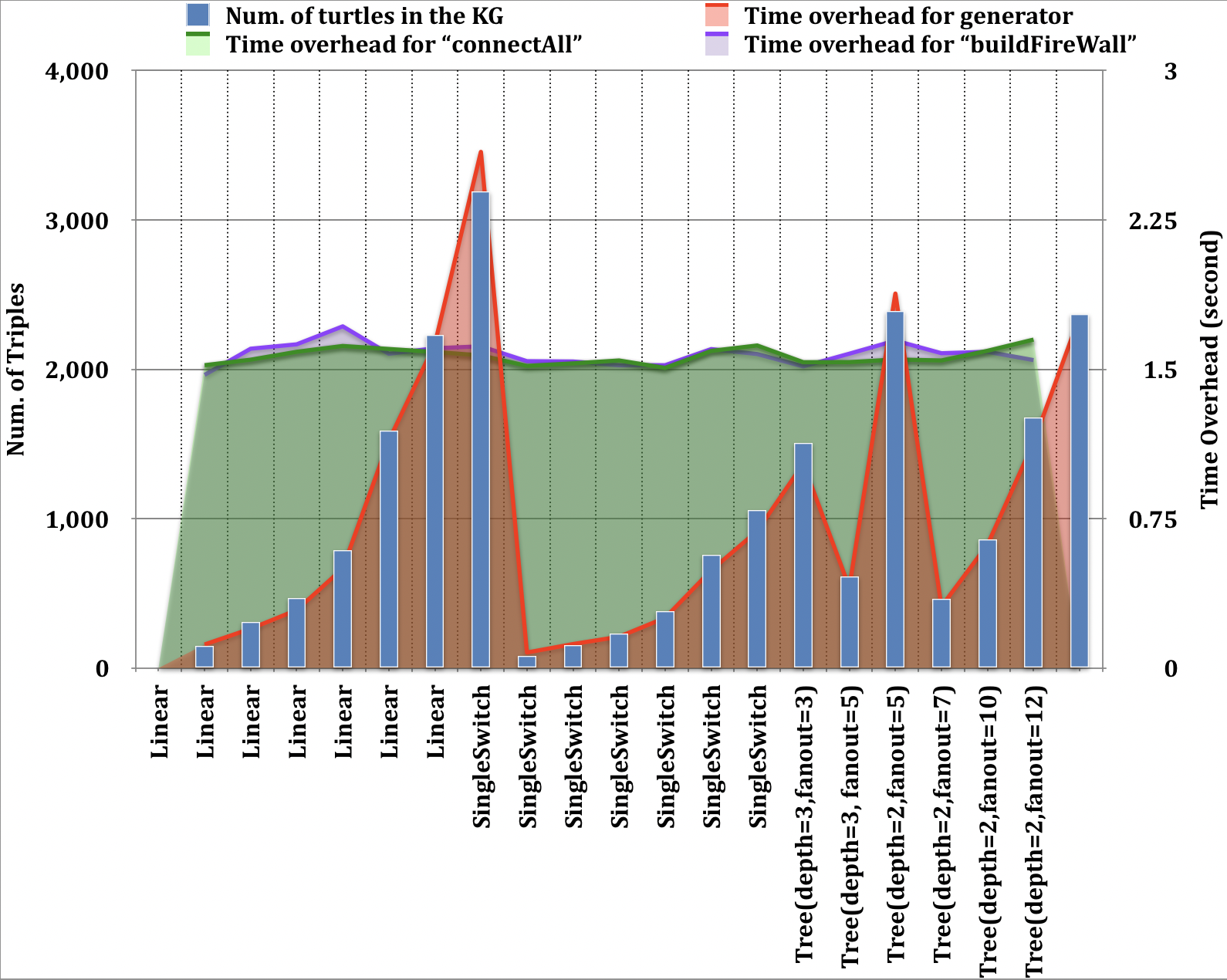}
	\caption{Execution time of knowledge graph constructor in computer networks with different scales, and time overhead of knowledge base generator and API method ``connectAll'' and ``buildFireWall''.}
	\label{fig_apps}
\end{center}
\end{figure}

\subsubsection{Evaluation of the knowledge base generator}
	Results of the experiments evaluating the knowledge base generator are presented in this section. As Ryu does not rely on ontological knowledge base, no comparison is made with it.
	\par
	In the first experiment, we try to evaluate the performance of the knowledge base generator in computer network with the metric of response time. We set up 19 networks with various scales, ranging from 6 nodes (single switch, \emph{k}=5) to 1146 nodes (tree topology, depth = 3, fanout = 5), and topologies of linear, single, and tree. The knowledge base generator is run on each network. For each run, we keep record of the start time, end time (defined by time point when the knowledge graph is completely written to the .rdf file), and the number of RDF triples of each knowledge graph generated. The results have been analyzed and compared using the metric of response time, as illustrated in Fig. \ref{fig_apps}. As shown in the figures, the response time grows linearly with the network scale. The network takes the slowest response with the most number of nodes. For the networks with different topologies but the same number of nodes, the simpler topology takes less time, e.g., the knowledge base generating speed for a simple switch network with 50 nodes is 0.65 milliseconds faster than the linear network with 50 nodes. Both the scale and the topology of the network affect the knowledge base generating efficiency. However, the overhead is acceptable for networks.
	
	In the second experiment, we evaluate the performance of the knowledge base generator in both computer network and WiFi network, with the metric of response time. We create 20 networks with 10 different topologies and scales (for each topology, e.g., linear topology with 5 hosts, one computer network and one WiFi network are generated). The knowledge base generator is executed on all the networks. In each run, the start time, end time and the number of RDF triples in the knowledge graph generated are recorded, and the results are shown in Fig. \ref{fig_apps}. The x-axis denotes the types of network topology, and the y-axis denotes the execution time overhead (defined by end time - start time). The performance results of WiFi networks are illustrated with purple circles, and computer networks with pink circles. The scales of networks are shown by the radius of the circles, varying from a small network with 5 nodes to a large one with more than 1000 nodes. The longest response time, 2.591602s, is taken by the largest network, which has more than 1000 nodes and links. As illustrated in Fig. \ref{fig_apps}, the response times of the knowledge base generated for both the computer and WiFi networks increase linearly with the increase in the number of RDF triples generated. However, the modeling processes for computer networks with different topologies and scales are generally slower than those of WiFi networks by marginal values. One possible reason is that it is more efficient to operate on WiFi networks than on computer networks in Mininet-WiFi. It is also apparent that the modeling process for networks with similar number of RDF triples (those dots with a similar sizes) but different topologies differ greatly. This is mainly due to the fact that the single switch topology is much simpler than the other topologies. Thus, it is the topology, rather than the number of RDF triples, that has a decisive effect on the modeling process for knowledge bases.

\subsubsection{Evaluation of the Network Management API}
	To evaluate the performance of the API, we executed the following methods to demonstrate how the basic network management tasks (such as adding a flow, listing all the flows of a switch) and complex tasks (such as connecting hosts to the network, and building a firewall) can be accomplished with a single line command. We compare the performance of SeaNet with Ryu -- a leading network management controller.
	
	\begin{description}
		\item[\emph{``add\_flow'':}] Add a flow entry to the flow table in a specific switch.
		\item[\emph{``dump\_all\_flows'':}] List all the flows in a switch.
		\item[\emph{``connectAll'':}] Connect all the hosts in a network automatically.
		\item[\emph{``buildFirewall'':}] Build a customised firewall between two or more switches. Block packets between certain hosts, while letting the other hosts under these switches pass.
	\end{description}
	
	\begin{table*}[ht]
		\vspace*{-\baselineskip}
		\caption{Comparison result of the SeaNet API ``connectAll'' and a ryu application with the same function}
		\label{table_ComparisonWithRyu}
		\begin{tabular}{|p{.15\textwidth}|p{.35\textwidth}|p{.4\textwidth}|}
			\hline
			\rowcolor[gray]{.7}
			& \Centering\textbf{Ryu App ``simple\_switch.py''} & \Centering\textbf{SeaNet API ``connectAll''}\\ \hline 
			\cellcolor[rgb]{.9,.9,.9}  Number of Lines in the source code & 120 lines & 5 lines for the query code\\  \hline
			\rowcolor[gray]{.9}
			\cellcolor[rgb]{.9,.9,.9} Code reusability & \textbf{Poor.} \par It is not an API, which means it is not reusable. & \textbf{High.} \par Plug-and-go \par and reusable.\\ \hline
			\cellcolor[rgb]{.9,.9,.9} Code Readability & \textbf{Poor.} Developers have to be quite familiar with Ryu architecture. & \textbf{Easy} \par API method call hides SPARQL query complexity \\ \hline
		\end{tabular}
		\vspace*{-\baselineskip}
	\end{table*}

	\begin{table*}\footnotesize
		\caption{The commands for building the same firewall with Ryu and SeaNet.}
		\label{table_comparasionComments}
		\begin{tabular}{|p{13cm}|p{3.3cm}|}
			\hline
			\rowcolor[gray]{.7}
			\Centering\textbf{Commands required by Ryu}&\Centering\textbf{Commands required by SeaNet}\\ [0.3em]\hline 
			\cellcolor[gray]{.9} \url{>} xterm c0& \\
			\cellcolor[gray]{.9} \url{#} \url{ovs-vsctl} set Bridge s1 \url{protocols=OpenFlow13} & \\
			\cellcolor[gray]{.9} \url{#} \url{ryu-manager} \url{ryu.app.rest_firewall} & \\
			\cellcolor[gray]{.9} \url{#} curl \url{-X} PUT \url{http://localhost:8080/firewall/module/enable/000000000001}&\\
			\cellcolor[gray]{.9} \url{#} curl \url{-X} POST $-d$ \url{'{"nw_src": "10.0.0.2/32", "nw_dst": "10.0.0.3/32"}'} \url{http://localhost:8080/firewall/rules/0000000000000001} & \url{#firewall(s1, s2, host1=h1, host2=h2)} \\
			\cellcolor[gray]{.9} \url{#} curl $-X$ POST $-d$ \url{'{"nw_src": "10.0.0.3/32", "nw_dst": "10.0.0.2/32"}'} \url{http://localhost:8080/firewall/rules/0000000000000001} & \\
			\cellcolor[gray]{.9} \url{#} curl $-X$ POST $-d$  \url{'{"nw_src": "10.0.0.2/32", "nw_dst": "10.0.0.3/32", "nw_proto": "ICMP", "actions": "DENY", "priority": "10"}'} \url{http://localhost:8080/firewall/rules/0000000000000001} &  \\ 
			\cellcolor[gray]{.9} \url{#} curl $-X$ POST $-d$  \url{'{"nw_src": "10.0.0.3/32", "nw_dst": "10.0.0.2/32", "nw_proto": "ICMP", "actions": "DENY", "priority": "10"}'} \url{http://localhost:8080/firewall/rules/0000000000000001} &\\
			\hline
		\end{tabular}
		\vspace*{-\baselineskip}
	\end{table*}
	
\begin{figure}[h]
		\vspace*{-\baselineskip}
		\begin{center}
			\includegraphics[width=3.3in]{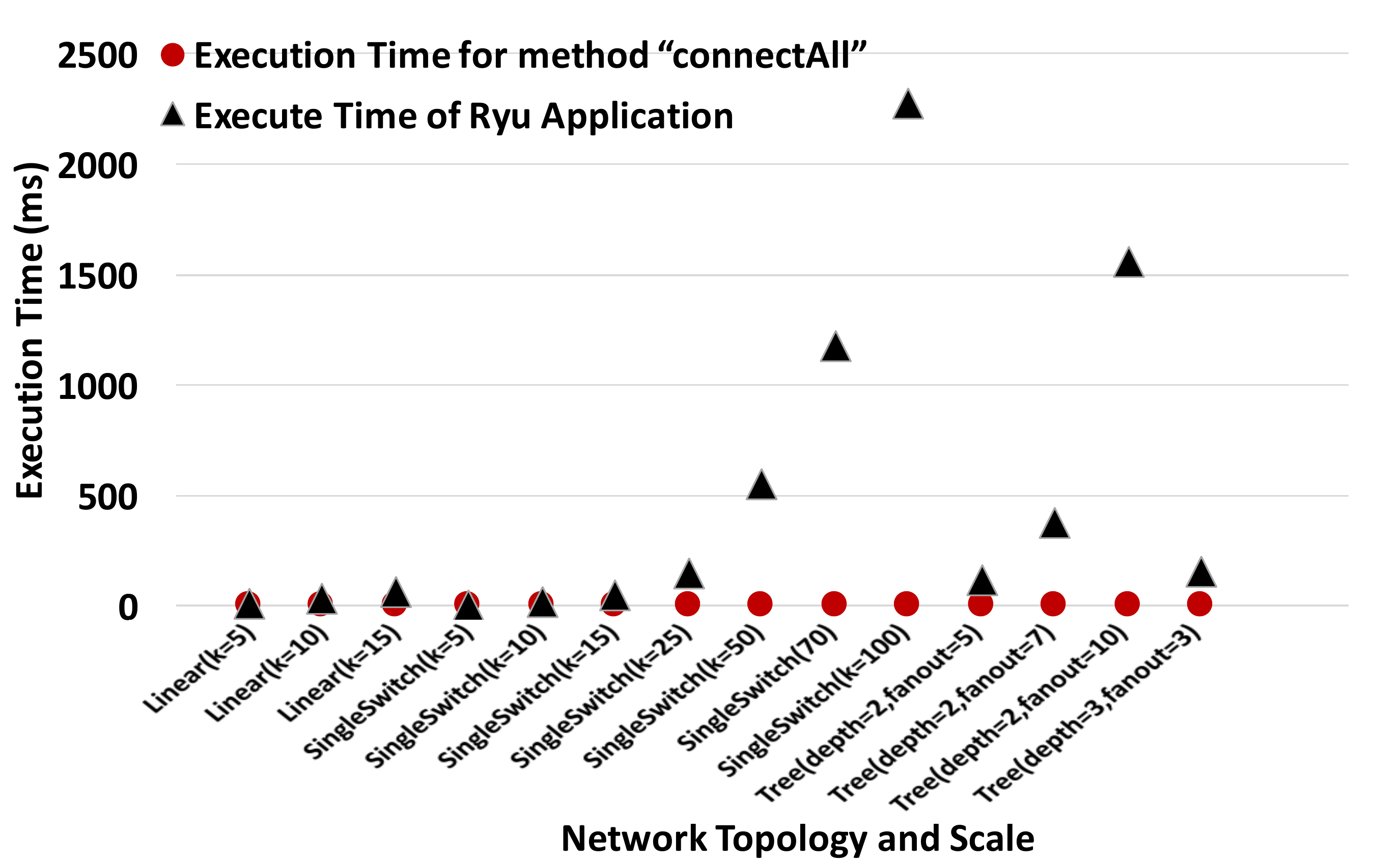}
			\caption{Execution time of Network Management API method ``connectAll'' and Ryu App with the same function, tested on computer networks with different topologies and scales. A black triangle represents the execution time of Ryu App, and the red dot represents the time of our API method ``connectAll'' on the same network.}
			\label{fig_comparasion}
		\end{center}
		\vspace*{-\baselineskip}
	\end{figure}

	In the first experiment, we set up a network with a switch ``s1'', and add a flow to switch ``s1'' with the API method \emph{``add\_flow''} and illustrated the results with the method \emph{``dump\_all\_flows''}. In the second experiment, 20 networks are emulated with various topologies and scales, from simple networks with 5 hosts to a large network with 1146 nodes. Two API methods \emph{``connectAll''} and \emph{``buildFirewall''}, and applications provided by Ryu, which accomplishes the same function as \emph{``connectAll''} and \emph{``buildFirewall''}, are executed on each of these networks. The experiment results are analyzed with the metrics of the time and operational efficiency, and code length. The results from our method \emph{``connectAll''} and Ryu application ``simple\_switch.py'', and our method \emph{``buildFirewall''} and the corresponding Ryu methods, are both compared. Evaluation results are shown in Figs. \ref{fig_apps}, \ref{fig_comparasion}, and Tables \ref{table_ComparisonWithRyu} and \ref{table_comparasionComments}. 
	
	In Fig. \ref{fig_comparasion} and Table \ref{table_ComparisonWithRyu}, the method \emph{``connectAll''} is compared with an existing application from Ryu.  With \emph{``connectAll''}, all the hosts inside the network are connected with each other, regardless of the network topology and the complexity. The application provided by Ryu, ``simple\_switch.py'', accomplishes the same function. The comparison results in terms of execution time on the same networks are shown in Fig. \ref{fig_comparasion}. Although the execution time of the Ryu application is equivalent to our method in small networks (with less than 15 nodes), it experiences immense increase as the network scale grows. The method \emph{``connectAll''}, however, has a stable performance. In large networks where the number of nodes is more than 100, the method \emph{``connectAll''} is almost 500 times more efficient than the corresponding application in Ryu. The code complexity is also compared between \emph{``connectAll''} and Ryu application, as shown in Table \ref{table_ComparisonWithRyu}, in terms of code length, code reusability, and code readability. It illustrated that our method saved up to $99.9\%$ execution time compared to to the corresponding Ryu application, with only $4\%$ of the code length and better readability and reusability.

	Method \emph{``connectAll''} was evaluated on computer network and WiFi networks, the time overhead is shown in Fig. \ref{fig_apps}, with the scalability of each knowledge graph generated. In Fig. \ref{fig_apps} the grey shadow in the background denotes the scale of each knowledge graph, while the time overhead of WiFi and computer networks are denoted with color-coded bars in blue and green respectively. It is obvious that the execution time goes almost unrelated with the scale of knowledge graph generated both in WiFi and computer networks. Actually, the time overhead is almost unchanged no matter what the scale of the knowledge graph is. It stays less than one second in all networks with various scales and topologies.  
	
	We compared the complexity of the \emph{``buildFirewall''} method with a Ryu application completing the same task, building a customized firewall between any nodes; the results are listed in Table \ref{table_comparasionComments}. The commands in Ryu are error prone and require specific knowledge of the network and programming to operate, whereas \emph{``buildFirewall''} method provided by SeaNet API is simply one line and self-explanatory. Users do not have to be an expert in the network or programming to execute it. The contrasting of technology intensive and technology independent operation is particularly clear. 
	
	The scalability of proposed methods is discussed in Fig. \ref{fig_apps}. The scale of a knowledge base is measured by the number of RDF triples it contains. Two methods ``connectAll'' and ``buildFireWall'' are executed on knowledge bases ranging from less than 100 RDF triples to more than 3000. Their execution time is illustrated in Fig. \ref{fig_apps}. The unit of the y-axis is a millisecond. From the figure, the advantage of RDF triple based reasoning is evident. The efficiency of RDF reasoning is not affected by the scale of RDF triple stored. This is due to the nature of RDF based reasoning, which is based on set operations, basic operation like finding a value, set-based operation would have time-complexity of $O(1)$ which is not related to the scale of the knowledge base, while in traditional database which is based on list, the complexity is $O(n)$ (n is the number of records).

	\section{Conclusions}\label{sec_conclusion_section}
	
	In this paper, we have taken a humble step towards a knowledge graph driven solution for autonomic management -- SeaNet. The heart of this system is a knowledge base for SDNs that uses knowledge representation and reasoning technologies to make autonomic knowledge inference for network management tasks possible. SeaNet consists of three parts, namely the knowledge base generator, the SPARQL engine, and the API. The functions provided by SeaNet have been proven to automatically accomplish some basic yet labor-intensive and error-prone network management tasks and be scalable. Time overheads for all the functions presented are limited to one second on networks with a scale up to more than 1000 nodes. Evaluations have been carried out based on networks with different technologies, topologies and scales. The performance has been analyzed using the metrics of response time, operational efficiency, and code efficiency. It is evident that benefiting from RDF reasoning, SeaNet is able to achieve O(1) time complexity on arbitrary scale of the network while the traditional database query efficiency can achieve O(nlogn) at its best. 
	

\bibliographystyle{IEEEtran} 
\bibliography{bare_jrnl_new_sample4}


\section{Biography Section}
\vspace{-33pt}
\begin{IEEEbiography}[{\includegraphics[width=1in,height=1.25in,clip,keepaspectratio]{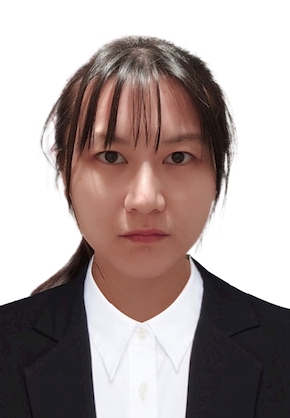}}]{Qianru Zhou} received her Ph.D. in electrical engineering from Heriot-Watt University, Edinburgh, U.K. in 2018. She is currently an Associate Professor in Nanjing University of Science and Technology, Nanjing, China. Her current research interests include declarative Artificial Intelligence, first order logic reasoning, zero-touch network management. 
\end{IEEEbiography}
\vspace{-33pt}
\begin{IEEEbiography}[{\includegraphics[width=1in,height=1.25in,clip,keepaspectratio]{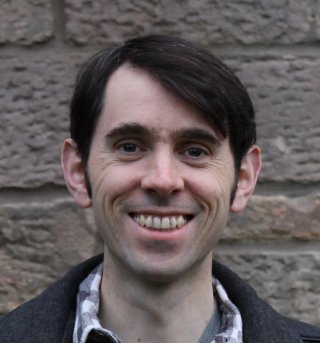}}]{Alasdair J.G. Gray} received the Ph.D. degree from Heriot-Watt University, U.K. in 2007. From 2013 he became a Lecturer in Computer Science at Heriot-Watt University. He held a postdoc research position at the University of Manchester between 2009 and 2013. Prior to that he held a postdoc research position at the University of Glasgow. His research interests focus on practical data management and its application in information systems ? utilizing and extending advances in knowledge management technologies to improve information systems. 
\end{IEEEbiography}
\vspace{-33pt}
\begin{IEEEbiography}[{\includegraphics[width=1in,height=1.25in,clip,keepaspectratio]{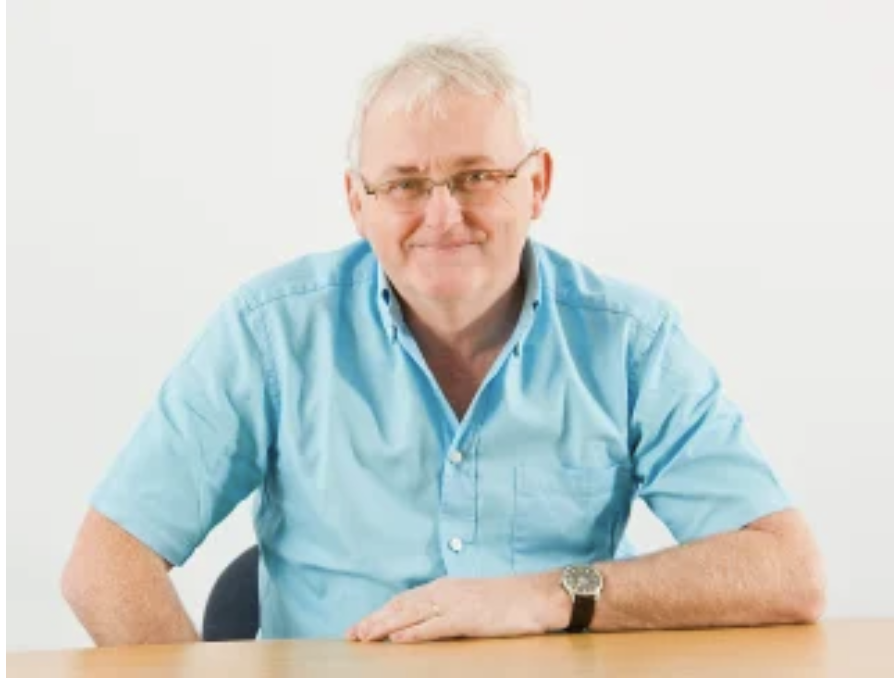}}]{Stephen McLaughlin}received the B.Sc. degree in Electronics and Electrical Engineering from the University of Glasgow in 1981 and the Ph.D. degree from the University of Edinburgh in 1989. He held a number of academic positions within the University of Edinburgh including that of Director of Research and Deputy Head of the School of Engineering. He has been the Head of School of Engineering and Physical Sciences at Heriot-Watt University since 2011. He is a Fellow of the Royal Academy of Engineering, Royal Society of Edinburgh, IET, and IEEE. 
\end{IEEEbiography}
\vspace{-33pt}
\vfill

\end{document}